\documentclass[runningheads]{llncs}
\usepackage{graphicx}
\usepackage[dvipsnames]{xcolor}
\usepackage{bm}
\usepackage{booktabs}       
\usepackage{multirow}
\usepackage{hyperref}
\begin{document}
\title{INN: Inflated Neural Networks \\ for IPMN Diagnosis}
%
%
\author{Rodney LaLonde\inst{1} \and
Irene Tanner\inst{1} \and
Katerina Nikiforaki\inst{2} \and
Georgios Z. Papadakis\inst{2} \and
Pujan Kandel\inst{3} \and
Candice W. Bolan\inst{3} \and
Michael B. Wallace\inst{3} \and
Ulas Bagci\inst{1}}

\authorrunning{R. LaLonde et al.}

\institute{University of Central Florida (UCF), Orlando FL, USA \and
Foundation for Research and Technology Hellas (FORTH),\\
Heraklion, Crete, Greece. \and 
Mayo Clinic, Jacksonville, FL, USA.}
\maketitle              
%
\begin{abstract}
Intraductal papillary mucinous neoplasm (IPMN) is a precursor to pancreatic ductal adenocarcinoma. While over half of patients are diagnosed with pancreatic cancer at a distant stage, patients who are diagnosed early enjoy a much higher 5-year survival rate of $34$\% compared to $3$\% in the former; hence, early diagnosis is key. Unique challenges in the medical imaging domain such as extremely limited annotated data sets and typically large 3D volumetric data have made it difficult for  deep learning to secure a strong foothold. In this work, we construct two novel ``inflated'' deep network architectures, \textit{InceptINN} and \textit{DenseINN}, for the task of diagnosing IPMN from multisequence (T1 and T2) MRI. These networks inflate their 2D layers to 3D and bootstrap weights from their 2D counterparts (Inceptionv3 and DenseNet121 respectively) trained on ImageNet to the new 3D kernels. We also extend the inflation process by further expanding the pre-trained kernels to handle any number of input modalities and different fusion strategies. This is one of the first studies to train an end-to-end deep network on multisequence MRI for IPMN diagnosis, and shows that our proposed novel inflated network architectures are able to handle the extremely limited training data (139 MRI scans), while providing an absolute improvement of $\bm{8.76}$\% in accuracy for diagnosing IPMN over the current state-of-the-art. Code is publicly available at \href{https://github.com/lalonderodney/INN-Inflated-Neural-Nets}{https://github.com/lalonderodney/INN-Inflated-Neural-Nets}.

\end{abstract}

\section{Introduction} \label{sec:intro}
\vspace{-0.2cm}
Pancreatic cancer is currently the fourth leading cause of cancer-related death in the United States in both men and women, behind Lung \& Bronchus, Breast (women)/Prostate (men), and Colon \& Rectum. These latter four forms of cancer have all seen massive declines in mortality over the past decades; however, the same cannot be said for pancreatic cancer, which has seen an increase in both occurrence and mortality from 2006 to 2017. Worse still, pancreatic cancer continues to carry one of the poorest prognoses of any form of cancer at a grim $9\%$ 5-year survival rate \cite{american_cancer_society_2019}. 
Because the cancer has usually spread beyond the pancreas by the time it is diagnosed, less than $20\%$ of patients are candidates for surgery. However, for cases in which diagnosis occurs while the disease is still local, which is true for approximately $10\%$ of patients, the 5-year survival rate has been steadily increasing from $29\%$ to $32\%$ to $34\%$ from 2017 to 2019 \cite{american_cancer_society_2019}. For these reasons, early diagnosis will be vital to increase the five-year survival rate of pancreatic cancer and make surgery a viable option for more patients.

\begin{figure}[htbp]
\vspace{-0.4cm}
\centering
\includegraphics[width=0.75\textwidth]{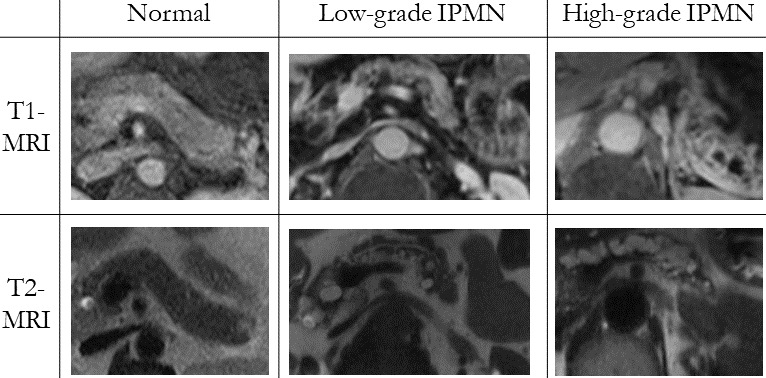}
\caption{Pancreas-ROIs from T1 \& T2-weighted MRI scans with the associated grades of IPMN from the post-surgery pathology report.} \label{fig:mri_roi}
\vspace{-0.8cm}
\end{figure}

\vspace{0.2cm}
\noindent\textbf{Diagnosing Intraductal Papillary Mucinous Neoplasm}\\ 
Intraductal papillary mucinous neoplasm (IPMN) is a radiographically detectable neoplasm that is found in the main and branch pancreatic ducts and is often a precursor to pancreatic ductal adenocarcinoma. IPMN has the potential to progress into invasive carcinoma, as a large proportion of main duct-IPMN exhibits malignant progression. Example magnetic resonance imaging (MRI) scans with the associated grade of IPMN are shown in Figure~\ref{fig:mri_roi}, where the scans are cropped to a regions of interest (ROI) surrounding the pancreas. In this study, the cases of IPMN were graded in a pathology report after surgery: 0) normal, 1) low-grade IPMN, 2) high-grade IPMN, and 3) invasive carcinoma. Being able to accurately detect IPMN early may help diagnose pancreatic cancer sooner and, in turn, increase the abysmally low 5-year survival rates. While progress has been made in diagnosing and managing IPMN pre-operatively using the radiographic criteria and international consensus guidelines, automated computer-aided diagnosis (CAD) systems are a strong candidate for helping with this task. CAD systems could be used to distinguish IPMN in the pancreas and differentiate between grades potentially more efficiently and effectively than current standards. 

\vspace{0.2cm}
\noindent\textbf{Literature on Classification of Pancreatic Cysts}\\
There are very few studies attempting to diagnose IPMN from MRI scans, and the presented work in this study is one of the first in literature to diagnose IPMN in MRI scans using deep learning. Sarfaraz et al. (2018) \cite{hussein_ipmn} introduced an architecture for automatic IPMN classification using a pre-trained 3D convolutional neural network (CNN) to perform feature extraction with canonical correlation analysis and feature fusion. Unlike \cite{hussein_ipmn}, our proposed approach allows for novel 3D network structures, intermediate fusion strategies, and end-to-end training. 

Although the goal is not related to IPMN diagnosis, Chen et al. (2018) \cite{chen_pcn} proposed PCN-Net, a study which can be considered a relevant work to ours because PCN-Net aims to classify pancreatic cystic neoplasms (PCNs). Briefly, the proposed network structure uses the 2D Inceptionv3  \cite{szegedy_inceptionv3} as a backbone, pre-trained on ImageNet \cite{russakovsky_imagenet} and fine-tuned on optical coherence tomography images. MRI sequences are aligned with a Z-Continuity Filter and combined using a late fusion strategy. Unlike \cite{chen_pcn}, our proposed approach utilizes 3D networks, again allows for intermediate fusion strategies, and does not require additional costly pre-training on external medical datasets. Furthermore, we focus on a relatively more challenging problem of IPMN diagnosis, not PCNs. Other studies in literature have focused on the use of pre-deep learning strategies such as conventional radiomics measurements (texture, intensity, and component enhancing features) to classify IPMNs. Not only is the success of these features fundamentally limited compared to deep learning-based approaches, but also these studies have focused on using CT images and require a priori segmentation of the pancreas \cite{gazit_ipmn}\cite{hanania_ipmn}, which itself is a challenging process.

\vspace{0.2cm}
\noindent\textbf{Proposed Solution: Multimodal Inflated Neural Networks}\\
While newer and more powerful deep network architectures continue to emerge, a difficulty arises when the new architecture differs in structure, preventing efficient weight transfer from pre-trained models. This is especially an issue in the medical imaging domain where annotated data is extremely limited and utilizing pre-trained deep networks often leads to dramatic performance increases. Additionally, due to the 3D nature of most medical imaging modalities, 3D networks tend to outperform their 2D counterparts, but most pre-trained networks exist only in 2D versions. There currently exists two solutions to this challenging problem. First, researchers can take their novel architecture and train on existing large datasets (e.g. ImageNet). This solution, however, is impractical and prohibitively expensive for many researchers. While some recent works in the literature have been proposed for shortening the typically multi-week process of training on ImageNet \cite{goyal_imagenet1hour}, these approaches typically require large clusters of GPUs for processing. 

The other solution is to cannibalize existing pre-trained networks and transfer weights in a process often referred to as ``network surgery''. Carreira and Zisserman \cite{carreira_i3d} used this strategy to great effect in a process they called ``network inflation'' to inflate the Inceptionv1 network for the task of action recognition. The network inflation process swaps out all 2D convolutional kernels and pooling operations for their 3D counterparts and replicates the kernels along the third dimension, while dividing the value of the weights by the number of replications to preserve relatively similar activation values. Liu et al. (2018) \cite{carreira_i3d} proposed AH-Net, which inflated part of its architecture from 2D for lesion segmentation. In this work, we follow a similar strategy of network inflation to inflate deeper, more advanced, and more complicatedly-connected networks while also tackling the issue of transferring weights when we have multiple imaging modalities. 

Several different fusion strategies exist for combining imaging modalities: early (pixel-level) fusion, intermediate fusion, and late fusion. In early fusion, images from different modalities are simply concatenated at the pixel level before being input to the network. In late fusion, inputs are fed to the network separately and the final embeddings are concatenated and fed to the final classification layers. Intermediate fusion is between early and late fusion, where information from multiple modalities are combined somewhere in the network. Which of these strategies is best depends on how similar the information is in each modality. Working with multiple modalities as input to a neural network adds an additional challenge for transferring weights from pre-trained networks, which typically accept only three-channel images as input. Transferring the weights of these earliest layers is critical for transfer learning, as these lowest-level layers have been found to share filters across virtually any imaging data, while higher-level layers become more specialized to the specific training data.

\vspace{0.2cm}
\noindent\textbf{Summary of Our Contributions}\\
\textbf{(1)} In this work, we construct two novel inflated deep network architectures, \textit{InceptINN} and \textit{DenseINN}, transferring weights from their 2D counterparts (Inceptionv3 \cite{szegedy_inceptionv3} and DenseNet121 \cite{huang_densenet} respectively) trained on ImageNet to the new 3D kernels. \textbf{(2)} We extend the inflation process by further expanding the pre-trained kernels to handle any number of input modalities and different fusion strategies. In general, this modification is the key component for any image-based diagnosis system accepting multiple image modalities as input. Particularly for our study, since T1- and T2-weighted MRI scans are the modality of choice for IPMN diagnosis, there is a strong need to combine the complementary information of both imaging modalities. The proposed network architecture provides the needed flexibility both at the input and fusion levels. \textbf{(3)} We investigate both early fusion as well as a version of intermediate fusion in greater detail and perform image-based diagnosis experiments at both the whole-MRI and pancreas-ROI levels for comprehensive evaluations, which have never been done in the literature before. \textbf{(4)} To our best of knowledge, our study is also the first one to train an end-to-end deep network on multisequence MRI for IPMN diagnosis. \textbf{(5)} Finally, we demonstrate that our newly designed inflated network architectures have a major advantage over the state-of-the-art by utilizing the extremely limited training data (139 MRI scans) while still obtaining an absolute improvement in accuracy for diagnosing IPMN of $\bm{8.76}$\%. 

\begin{figure}[htbp]
\centering
\includegraphics[width=0.9\textwidth]{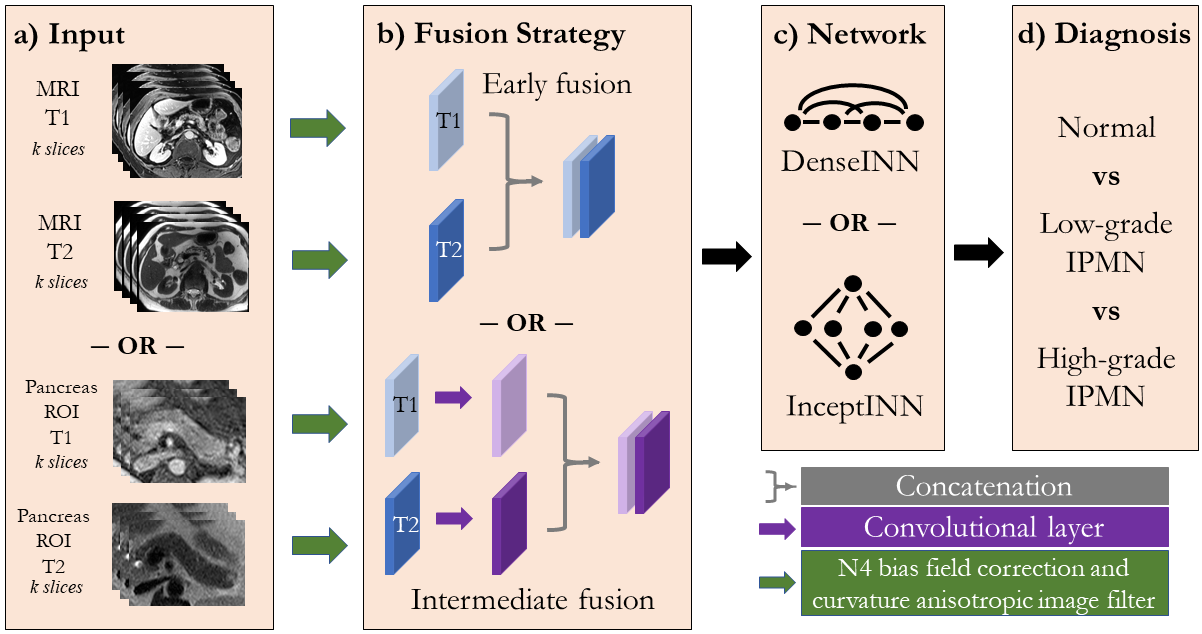}
\caption{Proposed network inflation framework overview.} \label{fig:framework}
\vspace{-0.55cm}
\end{figure}

\vspace{-0.4cm}
\section{Methods} \label{sec:methods}
\vspace{-0.2cm}
The overview of our proposed framework is shown in Fig.~\ref{fig:framework}. During training and testing, $k$ slices from either whole MRI scans or cropped pancreas-ROIs are used as input to one of our deep networks. Prior to entering the network, MRI scans are preprocessed by first aligning the T1 scans to the T2 scans using b-spline registration, then performing N4 bias field correction and applying a curvature anisotropic image filter. Following this, the $k$ slices of each modality are either concatenated along the channel axis ($height \times width \times slices \times channels$) and input to the network or are input separately, depending on if the early or intermediate fusion strategy is being followed. If intermediate fusion is chosen, each modality is passed through its own convolutional layer before being concatenated and passed through the remainder or the network. At the end of the network, either \textit{InceptINN} or \textit{DenseINN}, a softmax output over three values is obtained, representing the probability of our three possible grades of IPMN.

\vspace{0.2cm}
\noindent\textbf{INN: Inflating Neural Networks}\\ 
INN is a relatively straightforward two-step process. A 2D network is chosen for which pre-trained weights are available without needing major modifications to the overall network architecture that cannot be corrected with basic network surgery. The first step simply requires changing all 2D layers to their corresponding 3D counterparts (e.g. 2D convolution becomes 3D convolution, 2D pooling becomes 3D pooling). The choice of convolutional kernel size, pooling size, or stride length in this new third dimension is somewhat application dependant. For example, if training on a 30 fps video, one can afford larger kernels, strides, and pooling in the third dimension. For the application of IPMN diagnosis in MRI, since the pancreas typically only occupies a relatively small number of slices, we chose to favor strides of one in this new third dimension. For square kernels, we chose to extend these to cubes; however, for the $h \times 1$ or $1 \times w$ kernels, we chose to extend the third dimension to be $h \times 1 \times 1$ or $1 \times w \times 1$. 

The second step is to transfer the weights from the 2D kernels to the 3D kernels by bootstrapping them along the third dimension. This is accomplished by tiling the weights along the new dimension, then dividing all of the values of the kernel by the new depth of the kernel. This division is important to keep the network activations approximately the same from layer to layer. It may be straightforward when working with RGB images, but it can quickly become complicated with non-three-channel images and multiple imaging modalities.

\vspace{0.2cm}
\noindent\textbf{Handling Multiple Imaging Modalities}\\ 
There is a critical limitation of not having the proper number of kernels in the first convolutional layer when using a 2D network pre-trained on ImageNet with non-RGB images.
This becomes even more worrisome when you want to extend to multiple imaging modalities. In this work, we introduce the following strategies for handling multiple modalities while still allowing for the transfer of pre-trained weights. First, since the majority of medical imaging modalities (e.g. MRI, computed tomography) are single-channel, we tile the images to create three-channel images. This would be sufficient for a single modality input, but not for multiple modalities. If we are following the early fusion strategy, these three-channel images are concatenated prior to input into the network. Therefore, for each modality $M$, we create a copy of the first layer's kernels not only along the third dimension during the inflation process, but also along the input channel axis for the number of modalities given ($kern_h \times kern_w \times kern_d * 1 \times |M| * channels)$. This also increases the amount we divide the kernel values by a factor of $|M|$. 

If we are following the intermediate fusion strategy, the original first convolutional layer simply transfers copies of its weights (now also tiled along the new third dimension) to each modalities' individual convolution layer. These initial layer kernels are only divided by the length of the new third dimension. After these layers, the results are all concatenated and fed into the remainder of the network. The first layer after concatenation is now the one which must have $M$ copies made of its kernels along the input channel dimension and its values divided by $|M|$. \textit{DenseINN} adds some further complications due to the addition of concatenation layers starting from the second layer, but careful bookkeeping avoids any confusion.

\begin{table}[t]
\centering
\caption{Experimental results for IPMN diagnosis. Precision, recall, and accuracy are shown with the standard error across the ten-fold cross-validation splits. The proposed \textit{INN} networks outperforms their 3D trained from scratch versions (baselines) and the previous state-of-the-art at both the pancreas-ROI and whole-MRI levels.}
\label{table:results}
\begin{tabular}{@{}cc|cccccc@{}}
\toprule
Method & & & Pre (SEM)\% & & Rec (SEM)\% & & Acc (SEM)\% \\ 
\midrule
Hussein et al. \cite{hussein_ipmn} & & & - & & - & & $64.67$ ($0.83$) \\
\midrule
Baseline \textit{InceptINN} Whole-MRI & & & $61.59$ ($5.83$) & & $58.18$ ($4.05$) & & $61.87$ ($2.83$) \\
\textit{InceptINN} Whole-MRI & & & $74.51$ ($4.70$) & & $\bm{71.24}$ ($4.39$) & & $73.38$ ($3.52$) \\
\midrule
Baseline \textit{InceptINN} Pancreas-ROI & & & $68.49$ ($2.63$) & & $69.18$ ($3.40$) & & $69.37$ ($3.11$) \\
\textit{InceptINN} Pancreas-ROI & & & $71.68$ ($2.25$) & & $70.11$ ($1.33$) & & $72.32$ ($0.97$) \\
\midrule
Baseline \textit{DenseINN} Pancreas-ROI & & & $66.81$ ($4.51$) & & $67.05$ ($3.06$) & & $67.16$ ($2.93$) \\
\textit{DenseINN} Pancreas-ROI & & & $\bm{78.20}$ ($4.17$) & & $69.09$ ($2.97$) & & $\bm{73.43}$ ($2.26$) \\
\bottomrule
\end{tabular}
\vspace{-0.5cm}
\end{table}

\vspace{-0.4cm}
\begin{table}[htbp]
\centering
\caption{Examining the role of different fusion strategies. Precision, recall, and accuracy are shown comparing early versus intermediate fusion on a single training fold.}
\label{table:fusion}
\begin{tabular}{@{}c|ccc|ccc@{}}
\toprule
\multirow{2}{*}{Method} &
    \multicolumn{3}{c}{Early Fusion} & \multicolumn{3}{c}{Intermediate Fusion}\\
 & Pre \% & Rec \% & Acc \% & Pre \% & Rec \% & Acc \% \\
\midrule
\textit{InceptINN} Whole-MRI & $69.44$ & $66.35$ & $73.33$ & $66.67$ & $59.68$ & $66.67$ \\
\textit{InceptINN} Pancreas-ROI & $79.29$ & $\bm{80.48}$ & $78.57$ & $70.77$ & $69.66$ & $75.00$ \\
\textit{DenseINN} Pancreas-ROI & $73.08$ & $73.08$ & $75.00$ & $\bm{88.10}$ & $75.21$ & $\bm{82.14}$ \\
\bottomrule
\end{tabular}
\vspace{-0.5cm}
\end{table}

\vspace{-0.4cm}
\section{Experiments and Results}
Images were split across three categories: normal (i.e. no IPMN was present), low-grade IPMN, and high-grade IPMN/invasive carcinoma. Whole-MRI experiments used 139 scans, of which 29 were normal, 45 were low-grade, and 65 were high-grade. Due to the challenging nature of cropping the pancreas-ROI, two sets of crops were extracted by two experts. In cases where an expert could not confidently extract a pancreas-ROI, this scan was skipped, yielding 271 pancreas-ROIs. Experiments were carried out corresponding to each of the `-or-'s in our framework overview shown in Fig.~\ref{fig:framework}. The results are summarized in Table~\ref{table:results} for the intermediate fusion strategy. For all experiments, slices is set to $k = 5$, excluding the experiments specifically examining the effect of $k$ shown in Table~\ref{table:slices}. Unless otherwise specified, all experiments were conducted with stratified 10-fold cross-validation. Although the main focus of this work is on inflated networks and not modality fusion strategies, a set of experiments was conducted across one training fold to determine the relative performance of the two fusion strategies, with results shown in Table~\ref{table:fusion}. Note, experiments were not performed with \textit{DenseINN} for whole-MRI due to memory constraints. Whole-MRI images are resized in-plane to $256 \times 256$ and pancreas-ROIs are resized to $128 \times 128$.

All training and testing was performed using Keras with TensorFlow on a single Titan-X GPU with 12G memory. The Adam optimizer was used with its default parameters, early stopping, and learning rate reduction by $0.05$ on loss plateau. At training, input batches are formed by first sliding through each set of $k$ slices containing the pancreas in a given scan, before moving on to the next scan. At testing, $k$ slices are chosen around a central slice, where this slice is determined as the one in which the pancreas appears the largest. For pancreas-ROI images, a batch size of 32 was used for both networks. Due to memory limitations, \textit{InceptINN} used a batch size of 16 when using the whole-MRI. The average test time for InceptINN per patient on a single Titan X GPU is 0.2 sec for whole-slide MRI and 0.05 seconds for pancreas-ROI of MRI.

\vspace{-0.4cm}
\begin{table}[htbp]
\centering
\caption{Examining the role of 3D context information. Precision, recall, and accuracy are shown for \textit{InceptINN} on a single training fold on both whole-MRI and pancreas-ROI inputs. Note: the whole-MRI level at $k = 7$ slices is unable to fit into GPU memory, even at a reduced batch size, and thus was not included in experiments.}
\label{table:slices}
\begin{tabular}{@{}c|ccc|ccc@{}}
\toprule
\multirow{2}{*}{$k$ slices} &
    \multicolumn{3}{c}{Whole-MRI} & \multicolumn{3}{c}{Pancreas-ROI} \\
 & Pre\% & Rec\% & Acc\% & Pre\% & Rec\% & Acc\% \\
\midrule
$k = 3$ & $57.22$ & $56.83$ & $60.00$ & $82.83$ & $\bm{83.76}$ & $\bm{85.71}$ \\
$k = 5$ & $\bm{66.67}$ & $\bm{59.68}$ & $\bm{66.67}$ & $70.77$ & $69.66$ & $75.00$ \\
$k = 7$ & - & - & - & $\bm{85.45}$ & $81.91$ & $82.14$ \\
\bottomrule
\end{tabular}
\vspace{-0.6cm}
\end{table}

\section{Discussion and Concluding Remarks}
In this work, we demonstrate the effectiveness of inflating deep CNNs for the task of IPMN diagnosis, and train two different end-to-end, novel network architectures on the extremely limited multisequence MRI data set. Our proposed networks, \textit{InceptINN} and \textit{DenseINN}, outperform the previous state-of-the-art by over $8\%$ when operating either on a cropped bounding box around the pancreas or when using the entire MRI. One fundamental advantage of the proposed approach is that it is flexible enough to adapt almost any network structures to volumetric data while any number of input modalities can be handled. We have also showed that while \textit{InceptINN} tended to favor the early fusion approach, \textit{DenseINN} favored the intermediate fusion approach instead. Although our observations were empirical, full cross validation experiments can be performed as an extension of this study to determine if this is a data driven or network structure driven preference. 
\vspace{-0.4cm}

%
%
%
%

\end{document}